\documentclass[letterpaper]{article} 
\usepackage{aaai25}  
\usepackage{times}  
\usepackage{helvet}  
\usepackage{courier}  
\usepackage[hyphens]{url}  
\usepackage{graphicx} 
\usepackage{natbib}  
\usepackage{caption} 
\usepackage{algorithm}
\usepackage{algorithmic}
\usepackage{soul,color}
\usepackage{amsmath}
\usepackage{amsfonts}

\usepackage{newfloat}
\usepackage{listings}
\DeclareCaptionStyle{ruled}{labelfont=normalfont,labelsep=colon,strut=off} 
\floatstyle{ruled}
\newfloat{listing}{tb}{lst}{}
\floatname{listing}{Listing}
\title{TEncDM: Understanding the Properties of the Diffusion Model \\ in the Space of Language Model Encodings}
\author {
    Alexander Shabalin\textsuperscript{\rm 1,\rm 5}\thanks{First two authors contributed equally.},
    Viacheslav Meshchaninov\textsuperscript{\rm 1}\footnotemark[1],
    Egor Chimbulatov\textsuperscript{\rm 1},
    Vladislav Lapikov\textsuperscript{\rm 1},
    Roman Kim\textsuperscript{\rm 1},
    Grigory Bartosh\textsuperscript{\rm 2},
    Dmitry Molchanov\textsuperscript{\rm 3},
    Sergey Markov\textsuperscript{\rm 4},
    Dmitry Vetrov\textsuperscript{\rm 5}
}
\affiliations {
    \textsuperscript{\rm 1}HSE University\\
    \textsuperscript{\rm 2}University of Amsterdam\\
    \textsuperscript{\rm 3}Independent Researcher\\
    \textsuperscript{\rm 4}SberDevices\\
    \textsuperscript{\rm 5}Constructor University\\
\{amshabalin, vmeshchaninov,
echimbulatov, vlapikov, rkim\}@hse.ru,
dmolch111@gmail.com, g.bartosh@uva.nl, sergei.markoff@gmail.com,
dvetrov@constructor.university
}

\begin{document}

\maketitle

\begin{abstract}
This paper presents the \textit{Text Encoding Diffusion Model} (\textbf{TEncDM}), a novel approach to diffusion modeling that operates in the space of pre-trained language model encodings. In contrast to traditionally used embeddings, encodings integrate contextual information. In our approach, we also employ a transformer-based decoder, specifically designed to incorporate context in the token prediction process. We conduct a comprehensive examination of the influence of the encoder, decoder, noise scheduler, and self-conditioning on zero-shot generation. Furthermore, we compare \textbf{TEncDM} with previous approaches on three conditional text generation tasks: QQP, XSum, and Wiki-Auto. The results show that \textbf{TEncDM} exhibits superior performance compared to existing non-autoregressive diffusion models. Our code is available at \texttt{https://github.com/M0RJIQUE/tencdm}.
\end{abstract}

%

\section{Introduction}

Autoregressive (AR) large language models such as GPT-4 \cite{gpt4} or Llama 3 \cite{llama3} are the current gold standard in the text generation problem. They are capable of creating high-quality and coherent texts that are practically indistinguishable from the human ones. However, the disadvantage of this approach is the inability of the model to correct its own mistakes made during left-to-right generation. If a mistake occurs, it may spoil the subsequent text. In addition, the autoregressive method of token generation slows down the inference process as it requires performing a single model evaluation for each new token.

Diffusion modeling is currently the state-of-the-art approach for data generation in image \cite{latentdiffusion, sdxl}, audio \cite{stable_audio} and video \cite{stable_video_diffusion} domains. Researchers now attempt to adapt it also for text generation \cite{diffusion-lm, diffuseq, mahabadi2023tess}. Diffusion models are a class of probabilistic generative models that are able to iteratively transfer noise to a representative sample of data. While some of the proposed text diffusion models are autoregressive \cite{ld4lg, planner}, the majority of them are not and, by design, they have several advantages over AR language models. First, being non-autoregressive (NAR) models, they generate all the tokens simultaneously and can adjust any part of the sequence during the generation process. They also can be faster than AR models because the number of neural function evaluations for diffusion models depends on the number of denoising iterations rather than the length of the sequence. And given the possibility of distillation of diffusion models \cite{dm_distil}, the number of iterations can be greatly reduced.

To date, a number of text diffusion models have been proposed, each based on substantially new ideas with little overlap with other methods. Some works replace Gaussian noise with categorical noise \cite{multinomial_diffusion, d3pm}, exploiting the discreteness of the text domain. Others train continuous diffusion on token embeddings \cite{diffusion-lm, genie, diffuseq}, on text latent representations reduced in size \cite{ld4lg, planner} or on the probability simplex \cite{mahabadi2023tess, han2022ssd}. There are also differences in the way diffusion outputs are decoded back into text. Diffusion models trained on embeddings round their predictions to the nearest embeddings, while those that utilize small latent spaces decode the predictions with an AR model. This suggests that the scientific community has not yet found the most robust design of a diffusion model for text. 

In this paper, we attempt to better understand the specifics of continuous text diffusion models trained in the latent space of token embeddings and identify best practices for their development. We argue that the use of raw embeddings as latents is suboptimal, and that this approach can be enhanced by first extracting context information from embeddings with a language model encoder and training a diffusion model in this latent space. We also investigate several diffusion components in detail: text decoding methods, diffusion model architecture, noise schedule, and self-conditioning \cite{self-cond}. As a result, we combine all our findings in a method called \textit{Text Encoding Diffusion Model} \textbf{(TEncDM)}. 

We compare our approach with other non-autoregressive diffusion models trained in the space of embeddings or encodings on three conditional text generation problems – \textbf{paraphrasing}, \textbf{summarization}, and \textbf{text simplification} – and show its superiority over other methods. The main contributions of this work are as follows:
\begin{itemize}
    \itemsep0em
    \item We propose a new text diffusion framework \textbf{TEncDM}, which trains the diffusion model in the latent space constructed by the outputs of pre-trained Transformer-based encoder.
    \item We evaluate the importance of the text decoder and conclude that its robustness to inaccuracies in the generated latents directly affects the generation quality. We propose a Transformer-based decoder and its training method that boosts the model performance.
    \item We analyse in detail the effect of self-conditioning and noise schedules on the denoising process and show their effect on the model quality.
\end{itemize}

\section{Problem Statement and Background}

\paragraph{Text generation problem.}
In the field of natural language processing, unconditional text generation is a task of sampling $y$ from the unknown distribution $p(y)$, where $y = [y_1, \dots, y_n]$ is a
sequence of tokens with variable length $n$. In conditional text generation the distribution of texts changes to $p(y | x)$, where $x$ is a condition variable. The goal is to generate a text, that satisfies this condition.

\paragraph{Gaussian diffusion models.} 

Gaussian diffusion models \cite{score-based} learn to sample data from an unknown distribution by gradually denoising random Gaussian noise. The training procedure is defined through a forward diffusion process that satisfies $q(z_t|z_0) = \mathcal{N}(\sqrt{\alpha_t} z_0, (1 - \alpha_t) \mathbf{I})$, where $\alpha_t \in [0, 1]$ is a predefined noise schedule, $t \in [0, 1]$ and $\alpha_t > \alpha_{t+\Delta t}$.
The denoising network (parameterized by $\theta$) is trained to reconstruct the original latent $z_0$ given the noisy latent $z_t$, as expressed in equation \ref{eq::diff_loss}.
\begin{align}
\mathcal{L}(\theta) = \mathop{\mathbb{E}}_{\varepsilon \sim \mathcal{N}(0, \mathbf{I}), t \sim U[0; 1]} [\|z_0 - \hat{z}_{\theta}(z_t, t)\|^2] \label{eq::diff_loss}
\end{align}
Sampling procedure starts from a pure Gaussian noise $z_T \sim \mathcal{N}(0, \mathbf{I})$ and utilizes the denoising network to iteratively generate latents $z_{t_{T-1}},..., z_{t_1}$, where $1 = t_T > ... > t_1 = 0$.

\paragraph{Diffusion models for text generation.}

The primary feature of the text domain is the discreteness of its samples. In order to train a diffusion model on them, they must first be translated into continuous space. Consequently, alongside the denoising model, the diffusion framework incorporates an \textit{encoder} that maps tokens into the continuous latents and a \textit{decoder} that performs the reverse operation, converting the generated latents into text.

\section{Related Work}

\paragraph{Embedding-based diffusion models.}

The majority of proposed text diffusion models use embeddings of tokens to construct the continuous latent space \cite{diffusion-lm, genie, sed, diffuseq, ar-diffusion}. At the inference stage, to convert the latent predictions into text, they map each latent vector to a token corresponding to the nearest embedding.


\paragraph{Self-Conditioning.}
Self-conditioning is a technique that significantly increases the performance of the text diffusion model \cite{self-cond, sed, ld4lg}. Usually the model is conditioned only on the latent variable $z_t$ and the current timestep $t$ as $\hat{z}_0^t = \hat{z}_\theta(z_t, t)$. Self-conditioning proposes to also condition the model on the estimation of data sample from the previous timestep during generation in order to improve the prediction at the current timestep, $\hat{z}_0^{t} = \hat{z}_\theta(z_t, t, \hat{z}_0^{t-1})$.

Although widely used, no analysis has been conducted to determine why this method is effective or how the generation process is altered by its application.

\paragraph{Noise scheduler.}
Noise scheduler is a key component of a diffusion model that controls the amount of noise added on each timestep. Previous research \cite{diffusion-lm, difformer, dinoiser} has highlighted that the standard noise schedulers used for image diffusion models are unsuitable for the textual domain. Due to the discrete nature of the texts, it is unlikely that an addition of a small amount of noise to a latent will change its nearest text in the latent space. Therefore, to increase the difficulty of the denoising task for the model, the mentioned works recommend adding more noise on iterations that are close to 0.

\section{Understanding Text Diffusion}

In this section, we present our findings on the components of the diffusion model, discuss their weaknesses and propose ways to enhance them.

\paragraph{Encodings are better than embeddings.}
Most diffusion models utilize token embeddings to map text into a continuous latent space. However, this approach is not optimal because the embeddings do not convey contextual information. This requires the diffusion model to independently search for it to retrieve ambiguous tokens. To simplify the task, instead of embeddings, we can use the final layer outputs of a pre-trained language model (e.g. BERT). They contain contextual information and, thus, should be more suitable for training the diffusion model. We refer to these outputs as \textit{encodings}.

Experimental results confirming our intuition are presented in Section \ref{sec::encoders}. It is worth noting that the use of encodings does not slow down the generation process, as we need to compute them only during the training. To improve the quality even further, it is possible to fine-tune the encoder, but we choose not to in order to avoid overcomplicating the approach. Investigation into fine-tuning is left for the future work.

\paragraph{Decoder is important.}
The purpose of the decoder in the diffusion model is to map the generated latents into text. Approaches that train diffusion in the space of token embeddings decode latents by rounding them to the nearest embeddings and selecting a corresponding token. However, the diffusion model may produce inaccurate latent samples due to accumulation of errors during the denoising process. Such inaccuracy might significantly spoil the text quality, so it would be wise to train a decoder that could improve it.

In the Section \ref{sec::decoders}, we compare different decoder designs and conclude that an advanced context-dependent decoder indeed improves the generation quality. 

\paragraph{Self-conditioning affects denoising dynamics.}
Self-conditioning improves sampling quality by conditioning the model on its previous prediction. However, the mechanics of self-conditioning are not fully understood yet. Our research demonstrates that the addition of self-conditioning increases the model's prediction confidence at each denoising timestep, resulting in a reduction in the required number of generation steps. Furthermore, the sample quality diminishes as the number of steps increases. We believe that a reason for this behaviour lies in a mismatch between the latents used at the training stage and those at the generation stage. 
We provide the evidence supporting our conclusions in Section \ref{sec::self_condition}, along with a comprehensive analysis of the model's behaviour with and without self-conditioning.

\paragraph{Diffusion needs even more noise.}
Following the recommendations of previous works \cite{diffusion-lm, ar-diffusion, dinoiser}, we used \textit{sqrt} noise scheduler that increases the amount of noise added to the diffusion model inputs during training beyond the amount of typically used \textit{cosine} noise scheduler \cite{han2022ssd, ld4lg, sed, planner}. However, our experiments led us to conclusion that encoding-based diffusion model requires even more noise for successful training. We hypothesize that this is due to the presence of contextual information in the encodings, which simplifies the denoising task.

In Section \ref{sec::noise_schedulers} of this study, we demonstrate that both commonly used \textit{cosine} and \textit{sqrt} noise schedules do not introduce a significant level of noise to the latent variables over a wide range of timesteps. As a result, the denoising task becomes too simple for the model, leading to a reduction in the effectiveness of the training signal.

\section{Methodology}

\begin{figure}[t]
\begin{center}
\centerline{\includegraphics[width=\columnwidth]{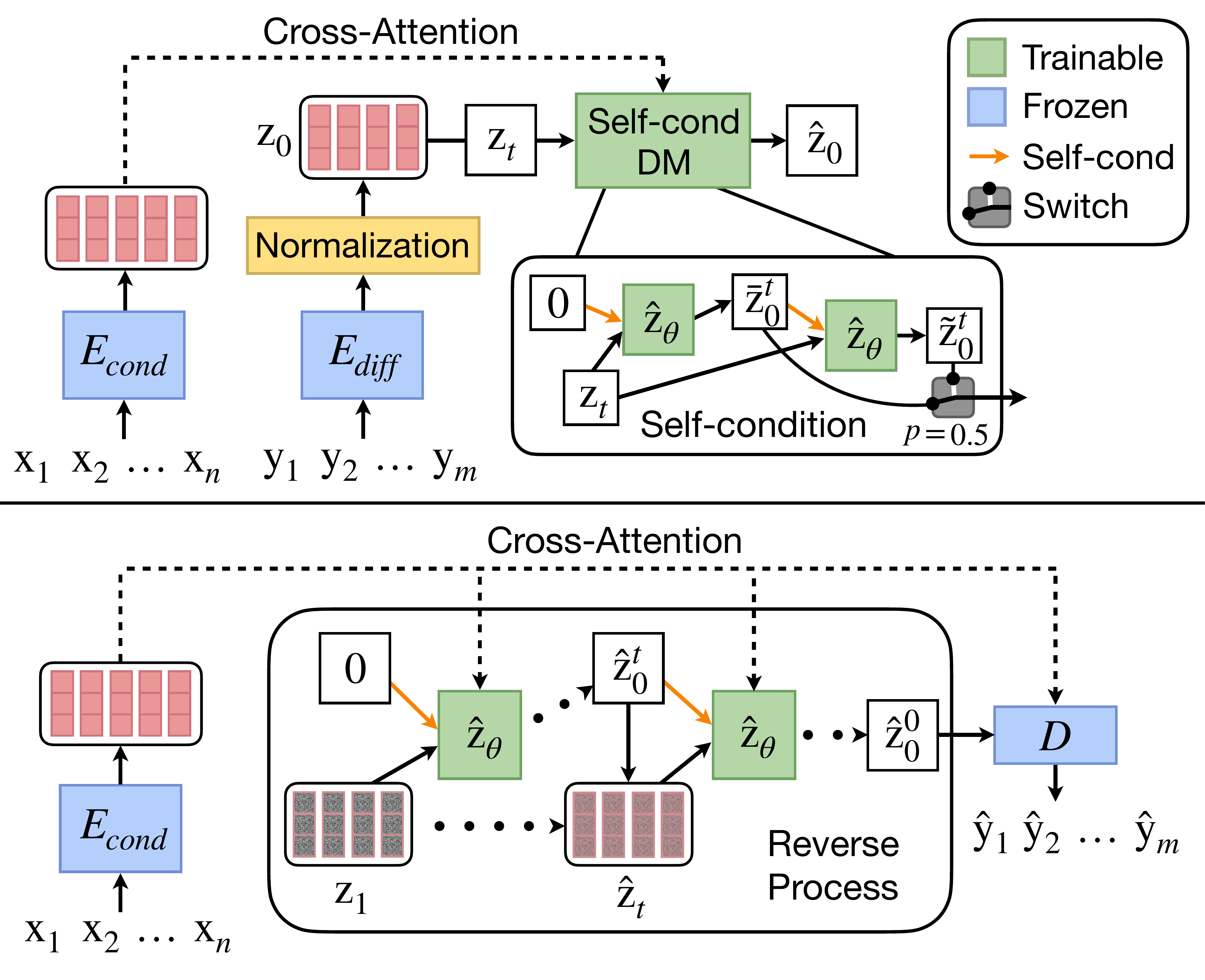}}
\caption{Overview of our framework design for conditional generation. Top is the training process, bottom is the generation process.}
\label{fig::model_pipeline}
\end{center}
\end{figure}

The design of \textbf{TEncDM} is depicted on Figure \ref{fig::model_pipeline}. It consists of three parts -- diffusion encoder $E_{diff}$, diffusion model $\hat{z}_\theta$ and decoder $D$. For the conditional generation, we also add condition encoder $E_{cond}$, which encodes an input text. Its output is provided to the diffusion model and decoder through cross-attention. 

This section exclusively focuses on the topic of unconditional text generation. The details of the conditional model can be found in Section \ref{sec::conditional}.

\subsection{Diffusion encoder, $E_{diff}$}

We use pre-trained Transformer-based \cite{transformer} language model $E_{diff}$, which we call \textit{diffusion encoder}, to encode text $y$ into the latent variable $z$. Encoding of text does not change the length of the sequence.
In order to align all texts in length, we add paddings to the end of short texts. After encoding the text, the encodings of all special tokens are replaced by their corresponding embeddings and padding encodings are replaced with zeros. This is necessary because diffusion model does not use an attention mask during training, which means that the reconstruction loss is calculated for both text and special tokens. However, special token encodings usually contain meaningless for diffusion model values. Therefore, minimization of reconstruction loss for these encodings only harms the training process. Embeddings of special tokens, on the other hand, only contain information about the token itself and the diffusion model recovers them much easier.
During training we do not update the weights of the encoder in order to keep the approach simple.

\subsection{Decoder, $D$}

The decoder $D$ is required to convert latent variables generated by diffusion model into textual output. Although a basic linear decoder can effectively reconstruct tokens with high accuracy, we employ the BERT-type \cite{bert} architecture for the decoder to provide it with the ability to capture context information and rectify potential mistakes originating from the diffusion model. Note that we do not use an AR decoder on purpose so as not to transfer the limitations of AR language models to the diffusion model.

We train the decoder independently of the diffusion model using the following objective
\begin{align}
-\mathbb{E}\log p_D(y \mid Cor(z_0)) \to \min_{D},
\end{align}
where $Cor(z_0)$ is a corrupted latent variable extracted from the diffusion encoder. Corruption is needed to expand the decoder training data domain and make it robust to distribution mismatch between text encodings $z_0$ and latents $\hat{z}_0$ generated by the diffusion model. This mismatch might arise due to the accumulation of errors during the denoising process. Its presence is especially evident for special tokens, which always have the same fixed representations in $z_0$. 
By default, we take $Cor(z_0)$ to be $z_t$ with randomly sampled $t \in [0, 0.15]$. We use the diffusion's noise scheduler to calculate $z_t$.

\subsection{Diffusion model, $\hat{z}_\theta$}

The diffusion model consists of 12 BERT layers and it is trained to predict the original latent $z_0$ given its noisy version $z_t$ and a timestep $t$ by minimizing the objective (\ref{eq::diff_loss}). We provide the model with timestep value by adding timestep embedding to the hidden state vectors on each layer.

We train the diffusion model using the variance preserving scheme, discussed in \cite{score-based}. To achieve zero mean and unit variance we normalize the latent variables $z_0$ coordinate-wise, using the statistics from the training set. 

\paragraph{Noise scheduler} We adopt the noise scheduler from \cite{simple_diffusion} and use the following equation for $\alpha_t$: 
\begin{align}
\alpha_t = \frac{1}{1 + \tan(t \pi / 2)^2 \cdot d^2},
\end{align}
where $d$ is a hyperparameter controlling the rate at which noise is introduced into the system. We set $d = 9$ by default, which corresponds to a significantly higher noise addition rate than what is used in all common noise schedulers. We further refer to our scheduler as \textit{tan-d} noise scheduler.


\paragraph{Self-condition} Following the previous approaches \cite{ld4lg, sed} we incorporate self-conditioning into the diffusion model. In order to make the model utilize the data sample estimation from the previous generation step, we modify the training procedure.

According to \cite{self-cond} we design the training process to emulate the inference behavior. On each training iteration with the probability $p = 0.5$ the prediction is computed with the self-conditioning set to zero $\bar{z}_0^t = z_\theta(z_t, t, 0)$. And, with probability $(1 - p) = 0.5$ we first calculate $\bar{z}_0^t = z_\theta(z_t, t, 0)$ and then use it as an estimation of the data sample to obtain a second prediction $\tilde{z}_0^t = z_\theta(z_t, t, \text{SG}(\bar{z}_0^t))$, where SG is the stop-gradient function that does not allow the gradient to flow through $\bar{z}_0^t$. The diffusion model is optimized using the output $\bar{z}_0^t$ in the former scenario and $\tilde{z}_0^t$ in the latter. This training strategy allows the model to accurately approximate $z_0$ both with and without self-conditioning. We implement self-conditioning in a same manner as conditioning on timestep. For each diffusion model layer  we pass the data estimation through a single linear layer and add it to the hidden state vectors.



\subsection{Generation process}
\label{sec::conditional}
The generation process is illustrated on the Figure \ref{fig::model_pipeline} (bottom). To generate text in the inference phase, we start with a random Gaussian sample and denoise it in $T$ steps using the Euler solver. At each step, we apply self-conditioning and, because of it, use a small number of steps -- $50$ by default. 


For the conditional generation we keep the framework design similar to unconditional generation. The only difference is that we add condition encoder to process the input text and provide both diffusion model and decoder with its output via cross-attention. We also add classifier-free guidance with coefficient 0.5, because it slightly improves quality. Implementation details can be found in Appendix E.

\section{Datasets}

To evaluate the performance of our diffusion models we use five datasets in English language. Two of them are unconditional: \textbf{ROCStories} and \textbf{Wikipedia}, and three are conditional: \textbf{QQP}, \textbf{XSum} and \textbf{Wiki-Auto}. The \textbf{ROCStories} \cite{rocstories} dataset contains 98k five-sentence commonsense fictional stories, that capture causal and temporal relations between daily events. The \textbf{Wikipedia} dataset is obtained from the ROOTS corpus \cite{bigscience}, it is a collection of over 2 million cleaned articles from the Wikipedia platform. The subset of \textbf{QQP} \cite{qqp} dataset, proposed in \cite{diffuseq}, consists of 144k question pairs from the Quora platform that are paraphrases of each other. The \textbf{XSum} \cite{xsum} dataset is used for summarization problem and it contains 204k BBC articles, which are provided as document and summary pairs. The \textbf{Wiki-Auto} \cite{wiki_auto} dataset consists of aligned sentences from complex and simplified Wikipedia\footnote[1]{All the datasets we use in this work are publicly available under a creative commons or an open source license.}. The detailed statistics for each dataset can be found in Appendix F.

\section{Empirical Analysis}

In this section, we evaluate the components of our framework on the \textbf{ROCStories} and \textbf{Wikipedia} datasets. To simplify the setup, we only consider unconditional generation. In Section \ref{sec::seq2seq}, we demonstrate that our findings can be successfully transferred to the conditional generation problems. In this section, we do not compare our method with others. The comparison with the GPT-2 \cite{gpt2} on unconditional generation is presented in Appendix J.

\subsection{Evaluation Metrics} 
We follow the model evaluation scheme from the \cite{ld4lg}. To evaluate the quality of our model we use \textbf{Perplexity (ppl)}, calculated with GPT-2 Large \cite{gpt2}. To measure the diversity of the generated text we utilize the diversity metric proposed in \cite{diversity}. We calculate it as \textbf{div}$(y) = \prod_{n=2}^4 \frac{|\text{\# of unique n-grams in } y|}{|\text{\# of n-grams in } y|}$, where $y$ is a set of generated texts. To ensure that the model does not reproduce the training dataset during the generation we evaluate the \textbf{Memorization (mem)}. We calculate it as the proportion of generated 4-grams that are found in the training set. As Perplexity tends to be small for the texts with repetitions, we also measure \textbf{MAUVE Score} \cite{mauve} to estimate the quality of text. MAUVE is a language model-based metric that measures the distance between the distributions of generated and reference texts using divergence frontiers. We leave all MAUVE hyperparameters at the default values presented in the original paper.

To calculate all the metrics, we generate 1000 texts. For MAUVE, we sample 1000 reference texts from the test set. We repeat this procedure 5 times and report the mean and standard deviation of the results in the $\text{mean}_{\text{std}}$ notation.

\subsection{Model setup}\label{sec::model_setup}

Unless otherwise stated, the training of \textbf{TEncDM} is performed within the latent space of BERT encodings. A three-layer transformer is employed for the decoder, which is trained to reconstruct $z_0$ from $z_t$, where $t \in U[0, 0.15]$. A comprehensive analysis of various decoder modifications is presented in this section and Appendix B.
The diffusion model is a 12-layer transformer with a dimensionality of 768. We train it with \textit{tan-9} noise scheduler.

\paragraph{Effect of Diffusion Encoder}\label{sec::encoders}

We compare latent spaces of BERT \cite{bert}, RoBERTa \cite{roberta}, BART \cite{bart} and T5 \cite{t5} encoders, as well as BERT embeddings, to ascertain the optimal choice for the diffusion model. All encoders have approximately the same size of 100M parameters. In this experiment, we train diffusion models with the same set of hyperparameters across all diffusion encoders. We train the decoders according to the scheme described in Section \ref{sec::model_setup}. The results of this comparison are presented in Table \ref{tab::encoders} and they show a clear advantage of the latent space derived from BERT encodings on ROCStories dataset. 
Furthermore, the quality of all encoders is superior to that of BERT embeddings. A better \textbf{div} and \textbf{mem} for embeddings can be explained by the presence of words in the corpus that do not align with the context. The text samples are presented in Appendix K.
This confirms our hypothesis that encodings are better suited for the training of a diffusion model. We discuss the drop in \textbf{mauve} for RoBERTa in Appendix G.

\begin{table}
\centering
\begin{tabular}{l|llll}
\hline
\multicolumn{1}{l|}{\textbf{Encoder}}
    & \multicolumn{1}{c}{\textbf{ppl} $\downarrow$} 
    & \multicolumn{1}{c}{\textbf{mem} $\downarrow$}
    & \multicolumn{1}{c}{\textbf{div} $\uparrow$}
    & \multicolumn{1}{c}{\textbf{mauve} $\uparrow$}\\
\hline
\multicolumn{5}{c}{\textbf{ROCStories}} \\
\hline
BERT emb & $48.9_{.36}$ & $.371_{.003}$ & $.324_{.002}$ & $.600_{.016}$ \\
BERT & $29.1_{.89}$ & ${.453}_{.003}$ & ${.295}_{.002}$ & $\textbf{.762}_{.043}$ \\
RoBERTa & $\textbf{28.3}_{.33}$ & ${.443}_{.003}$ & ${.302}_{.002}$ & ${.647}_{.019}$\\
T5 & ${31.3}_{.54}$ & $.427_{.003}$ & $.312_{.004}$ & $.706_{.024}$ \\
BART & $34.1_{.52}$ & $.441_{.006}$ & $.299_{.005}$ & $.705_{.030}$\\
\hline
Source text & $21.7$ & $.365$ & $.403$ & $.876$ \\
\hline
\multicolumn{5}{c}{\textbf{Wikipedia}} \\
\hline
BERT emb & $156.1_{1.8}$ & $.263_{.004}$ & $.517_{.002}$ & $.378_{.055}$ \\
BERT & $\textbf{104.4}_{2.1}$ & $.286_{.002}$ & $.504_{.003}$ & $\textbf{.874}_{.011}$ \\
\hline
Source text & $37.3$ & $.122$ & $.615$ & $.957$ \\
\hline
\end{tabular}
\caption{Quality of diffusion models trained with different diffusion encoders.}
\label{tab::encoders}
\end{table}

\paragraph{Effect of Decoder}\label{sec::decoders}

To confirm the hypothesis about the importance of the decoder architecture and its training scheme, we compare an MLP decoder consisting of two linear layers with a 3-layer transformer. The latter is able to extract contextual information for each token. We corrupt the decoder input $z_0$ by transforming it into $z_t$, using the diffusion forward process with $t \in U[0, 0.15]$. We choose this method, because it brings the decoder input closer to the diffusion output. A more detailed analysis of corruption techniques is presented in the Appendix B.
To keep the experiment fair, we apply all decoders to the same generated latents. The results of the experiment are shown in Table \ref{tab::decoders}. The MLP decoder without corruption achieves the lowest text quality in terms of \textbf{perplexity}, but comparable by \textbf{mauve} with Transformer without corruption. However, it is challenging to make meaningful comparisons between decoders by \textbf{mauve} due to the significant variance. From this experiment, we can conclude that corruption of the latent helps to improve the quality for both datasets. At the same time, the incorporation of contextual information into the decoder lead to the best result.

\begin{table}
\centering
\setlength{\tabcolsep}{1.7mm}
\begin{tabular}{l|cccc}
\hline
\textbf{Decoder} & \textbf{ppl} $\downarrow$ & \textbf{mem} $\downarrow$ & \textbf{div} $\uparrow$ & \textbf{mauve} $\uparrow$ \\
\hline
\multicolumn{5}{c}{\textbf{ROCStories}} \\
\hline
MLP & $39.7_{3.38}$ & $.444_{.002}$ & $.297_{.004}$ & $.716_{.074}$ \\
\; + $Cor(z_0)$ & $31.2_{.33}$ & $.448_{.002}$ & $.293_{.003}$ & $.739_{.051}$ \\
Transformer & $34.2_{.29}$ & $.445_{.001}$ & $.295_{.003}$ & $.714_{.037}$ \\
\; + $Cor(z_0)$ & $\textbf{29.1}_{.89}$ & ${.453}_{.003}$ & ${.295}_{.002}$ & $\textbf{.762}_{.043}$ \\
\hline
\multicolumn{5}{c}{\textbf{Wikipedia}} \\
\hline
Transformer & $180.6_{3.2}$ & $.261_{.001}$ & $.511_{.001}$ & $.526_{.025}$ \\
\; + $Cor(z_0)$ & $\textbf{104.4}_{2.1}$ & $.286_{.002}$ & $.504_{.003}$ & $\textbf{.874}_{.011}$ \\
\hline
\end{tabular}
\caption{Comparison of decoders for encoding-based diffusion model.}
\label{tab::decoders}
\end{table}

\begin{figure}[t]
\begin{center}
\centerline{\includegraphics[width=0.85\columnwidth]{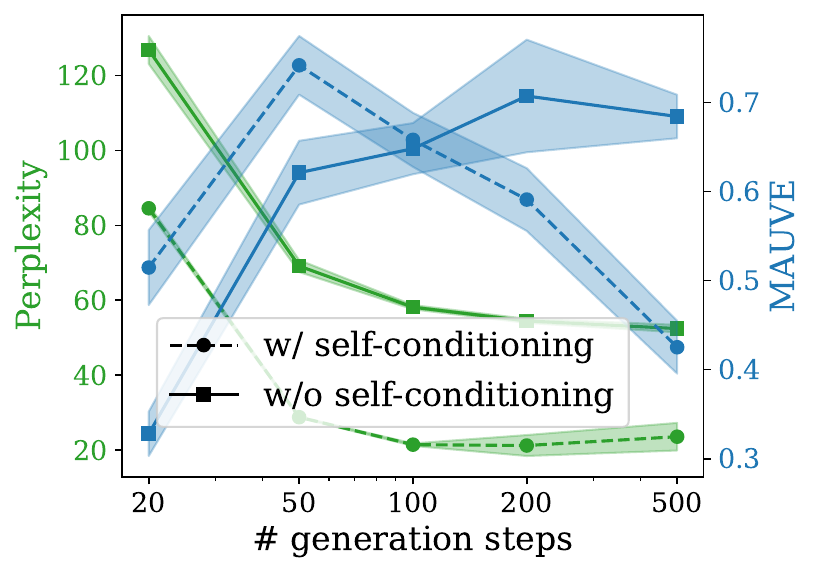}}
\caption{Generation quality of diffusion models with and without self-conditioning on ROCStories dataset.}
\label{fig::w_and_wo_sc}
\end{center}
\end{figure}

\paragraph{Effect of self-conditioning}\label{sec::self_condition}
We conduct a series of experiments to understand how self-conditioning (SC) affects the denoising process. In Figure \ref{fig::w_and_wo_sc}, we compare the quality of the models with and without SC for different number of denoising steps on the ROCStories dataset. The results show that while the quality of the model without SC increases as the number of steps increases, the quality of the model with SC reaches a maximum at a value of 50 steps in terms of \textbf{mauve}, and then it starts to drop. Nevertheless, at the highest point the model with SC surpasses the model without it according to both \textbf{mauve} and \textbf{perplexity}.

\begin{figure}[t]
\begin{center}
\centerline{\includegraphics[width=0.7\columnwidth]{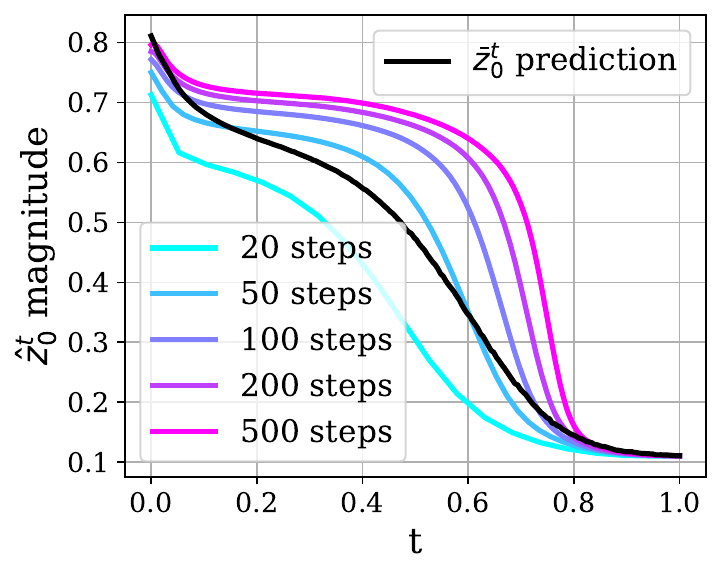}}
\caption{Prediction magnitudes for generation processes with different amount of steps on ROCStories dataset.}
\label{fig::z0_var_steps} 
\end{center}
\end{figure}

We explain this drop in generation quality with mismatch between diffusion model inputs at train and inference stages. 
To confirm our hypothesis, we calculated the mean-squared norm (\textit{magnitude}) of the values of each latent $\hat{z}_0^t$ in a mini-batch predicted by the diffusion model during generation (i.e. $\frac{1}{N \cdot d \cdot m}\|\hat{z}_0^t\|_2^2$, where $N$ is a batch size, $d$ is a dimension and $m$ is a sequence length). We plot this magnitude with respect to timestep for generations with different number of steps as well as for the predictions $\bar{z}_0^t$ from the training stage.
The results for the ROCStories dataset are presented in Figure \ref{fig::z0_var_steps}. They indicate that self-conditioning significantly increases the prediction magnitude as the number of steps increases.
This can be explained by the following: during training, the model learns to use self-conditioning to approximate $z_0$ more accurately. Consequently, self-conditioning increases the model's confidence, which is directly related to prediction magnitude. 
During the generation process, the model takes its own prediction, which has an increased magnitude, as an input at each step and increases it further. Therefore, the increase in magnitude depends directly on the number 
of generation steps. Eventually, this leads to a mismatch between the predictions fed into the model during training and generation. 

In the Appendix C, we provide a more detailed discussion of this phenomenon and show that the same behavior is observed in the larger Wikipedia dataset.
It is worth noting that the smallest mismatch is observed for the trajectory of 50 generation steps, which corresponds to the best quality.

\paragraph{Effect of Noise scheduler}\label{sec::noise_schedulers}

\begin{figure}[t]
\begin{center}
\centerline{\includegraphics[width=1\linewidth]{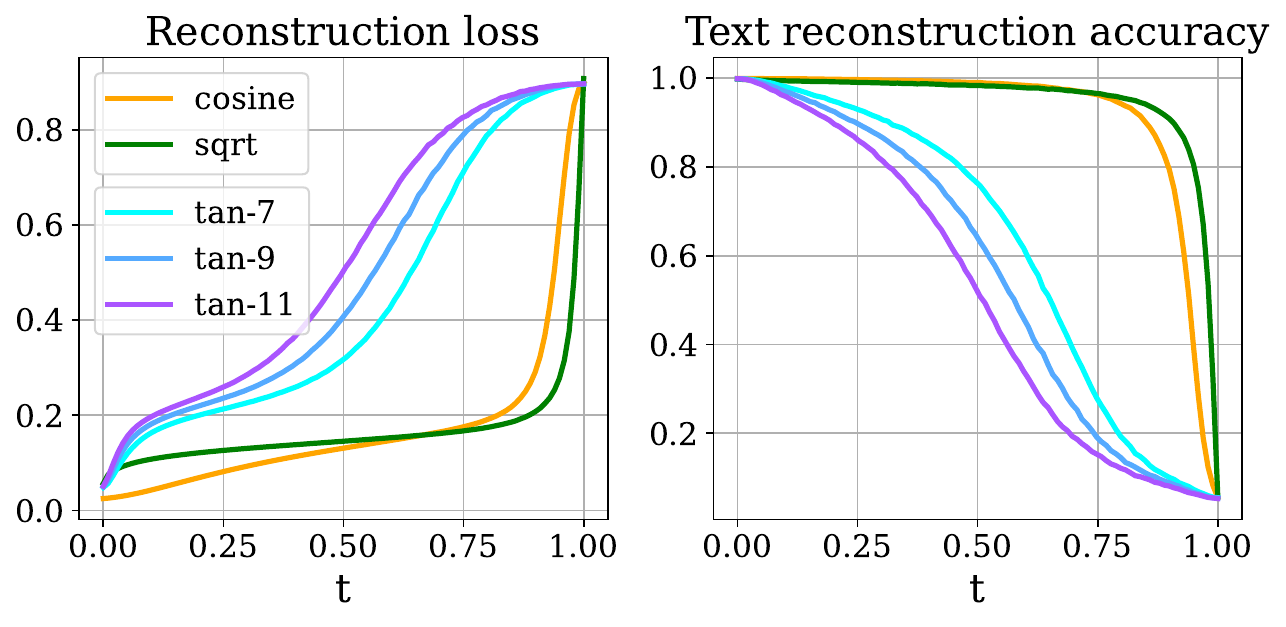}}
\caption{Reconstruction loss and reconstruction accuracy of diffusion models trained with different noise schedulers on ROCStories dataset.}
\label{fig::schedulers}
\end{center}
\end{figure}

We compare our noise scheduler \textit{tan-d} with previously used \textit{cosine} and \textit{sqrt} (visualized in Appendix D) and present the quantitative results in Table \ref{tab::schedulers}. We use the same decoder and optimal amount of generation steps for each scheduler.
In Figure \ref{fig::schedulers}, we evaluate the difficulty of recovering a data sample from noised latent $z_t$ for diffusion model trained with different noise schedulers. We measure the reconstruction loss $\frac{1}{N \cdot d \cdot m}\|z_0 - \bar{z}_0^t\|_2^2$ and accuracy of token prediction for every timestep.

While the \textit{sqrt} noise scheduler adds significantly larger amount of noise in the initial timesteps than \textit{cosine} one, the rate of noise addition decreases for the subsequent timesteps. For both schedulers, the denoising task becomes insufficiently hard for the most timesteps, which leads to a decrease in their contribution to the generation process. This can be seen from the reconstruction accuracy. In contrast, \textit{tan-d} noise scheduler adds more noise consistently across all timesteps, leading to a more challenging training task and improved generation performance.

Based on these observations, we conclude that in order to improve the efficiently of the denoising process, it is essential to increase the amount of added noise within all timesteps. However, it is important to strike a balance as adding excessive noise can negatively impact performance. In our experiments, \textit{tan-9} yielded results that were marginally superior to all other schedulers in terms of \textbf{mauve}, with slightly better \textbf{memorization} and \textbf{diversity}, while lagging behind \textit{tan-7} in terms of \textbf{perplexity}.

As a rule of thumb, the noise schedule should be such that the diffusion model recovers approximately the same amount of information at each timestep. Otherwise, some of them will not contribute to the denoising process enough.

\begin{table}
\centering
\small
\begin{tabular}{l|cccc}
\hline
\textbf{\begin{tabular}{@{}c@{}}Noise \\ Scheduler\end{tabular}} & \textbf{ppl} $\downarrow$ & \textbf{mem} $\downarrow$ & \textbf{div} $\uparrow$ & \textbf{mauve} $\uparrow$ \\
\hline
\multicolumn{5}{c}{\textbf{ROCStories}} \\
\hline
cosine & $393.2_{127.6}$ & $.262_{.004}$ & $.474_{.006}$ & $.098_{.011}$ \\
sqrt & $127.2_{29.3}$ & $.264_{.004}$ & $.434_{.004}$ & $.364_{.041}$ \\
\hline
tan-7 & $\textbf{27.1}_{.31}$ & $.455_{.004}$ & $.286_{.001}$ & $.730_{.026}$ \\
tan-9 & $29.1_{.89}$ & ${\textbf{.453}}_{.003}$ & ${\textbf{.295}}_{.002}$ & $\textbf{.762}_{.043}$ \\
tan-11 & $33.4_{2.4}$ & $.464_{.006}$ & $.279_{.004}$ & $.730_{.038}$ \\
\hline
\multicolumn{5}{c}{\textbf{Wikipedia}} \\
\hline
sqrt & $364.0_{6.5}$ & $.139_{.001}$ & $.664_{.004}$ & $.325_{.037}$ \\
tan-9 & $\textbf{104.4}_{2.1}$ & $.286_{.002}$ & $.504_{.003}$ & $\textbf{.874}_{.011}$ \\
\end{tabular}
\caption{Quality of diffusion models trained with different noise schedulers.}
\label{tab::schedulers}
\end{table}

\begin{table*}[t]
    \centering
    \begin{tabular}{l|ccc|cc|ccc}
\hline
    \textbf{Method} 
    & \multicolumn{3}{c}{QQP} 
    & \multicolumn{2}{c}{XSum} 
    & \multicolumn{3}{c}{Wiki-Auto} \\

    & \textbf{R-L} $\uparrow$         & \textbf{BS} $\uparrow$ & \textbf{B-4} $\uparrow$ 
    & \textbf{R-1/2/L} $\uparrow$     & \textbf{BS} $\uparrow$
    & \textbf{R-L} $\uparrow$         & \textbf{BS} $\uparrow$ & \textbf{B-4} $\uparrow$ 
    \\
\hline
    
\hline
    DiffuSeq$^\star$           & 52.7   & 82.4   & —    & 18.9 / 1.3  / 13.6  & 46.8 & —    & 79.1 & 26.1  \\
    SeqDiffuSeq$^{\star \dagger}$        & —      & 82.9   & 23.3 & 14.1 / 1.1  / 11.4  & 58.4 & —    & \textbf{82.1} & 37.1  \\
    GENIE$^\star$              & —      & —      & —    & 29.3 / 8.3  / 21.9  & —    & —    & —    & —     \\
    AR-Diffusion$^\dagger$       & 54.9   & 81.4   & \textbf{31.2} & 27.1 / 6.4  / 20.8  & 59.7 & 54.9 & 81.3 & 32.7 \\
    TEncDM (BERT)      & 56.4   & 82.4   & 30.2 & 31.5 / 10.0 / 24.9  & 68.2 & \textbf{58.1} & 80.5 & \textbf{41.6} \\
    TEncDM (T5)        & \textbf{57.3}   & \textbf{83.8}   & 30.7 & 33.4 / 11.4 / 26.8  & \textbf{70.1} & 57.7 & 81.2 & \textbf{41.6} \\
    TEncDM (RoBERTa)   & 55.8   & 82.4   & 30.0 & $\textbf{33.7}$ / $\textbf{11.9}$ / $\textbf{27.1}$ & 69.8 & 57.9 & 81.0 & 40.5 \\

\hline
    GPT2-small FT$^\star$    & $52.1$ & $82.5$ & $19.8$ & —                        & —      & $54.6$ & $80.2$ & $30.8$ \\
    Transformer-base$^\star$ & $57.5$ & $83.8$ & $27.2$ & $30.5$ / $10.4$ / $24.2$ & —      & $49.1$ & $73.8$ & $26.9$ \\
    FLAN-T5-base$^\star$     & $52.3$ & $83.2$ & —      & 34.6 / 12.9 / 27.2 & 72.7 & —      & —      & —      \\
\end{tabular}
\caption{
    Seq2Seq evaluation results of Diffusion and AR methods on QQP, XSum and Wiki-Auto datasets. 
    We calculate \textbf{ROUGE-1/2/L (R-1/2/L)}, \textbf{BERTScore (BS)} and \textbf{BLEU-4 (B-4)}. All results taken from other papers are marked with $\star$. DiffuSeq and SeqDiffuSeq results were taken from their respective publications. Results for AR models were taken from \cite{diffuseq, ar-diffusion, ld4lg}. Additionally, we trained AR-Diffusion and SeqDiffuSeq on previously unreported datasets, using the code from the corresponding papers (marked as $\dagger$).
}
\label{tab::tasks}
\end{table*}

\section{Seq2Seq Experiments}\label{sec::seq2seq}

We are conducting experiments to evaluate the effectiveness of the proposed components for text diffusion generation on language model encodings on three different tasks: paraphrasing (QQP), summarization (XSum), and text simplification (Wiki-Auto).

\paragraph{Baselines} 
We include two groups of baselines in comparison. The first group comprises of popular diffusion methods: DiffuSeq \cite{diffuseq}, SeqDiffuSeq \cite{yuan2022seqdiffuseq}, GENIE \cite{genie}, AR-diffusion \cite{ar-diffusion}. We focus only on non-autoregressive diffusion models trained in the latent space of embeddings or encodings. Besides, we compare TEncDM to classical AR baselines: Transformer \cite{vaswani2017attention}, FLAN-T5-base \cite{chung2024scaling} and finetuned GPT-2-small \cite{gpt2}. 

\paragraph{Metrics} 
For evaluation of paraphrasing and simplification tasks, we adopt the setting of SeqDiffuSeq \cite{yuan2022seqdiffuseq} and calculate ROUGE-L \cite{lin2004rouge}, BERTScore \cite{zhang2019bertscore} and BLEU-4.
In addition, we follow the approach of \cite{ar-diffusion} and report ROUGE-1/2/L for summarization task.

\paragraph{Results} 
Table~\ref{tab::tasks} presents a comprehensive comparison of our approach against existing methods across three datasets. Results for DiffuSeq, and SeqDiffuSeq were sourced from their respective papers \cite{ar-diffusion, yuan2022seqdiffuseq}. Results for GENIE were taken from the \cite{ar-diffusion}, as the original paper used ground truth labels to select the best generated text, which introduces unfairness.
Additionally, we trained both AR-Diffusion and SeqDiffuSeq on previously unreported datasets, using the code from the corresponding papers.
Results for AR models were taken from \cite{diffuseq, ar-diffusion, ld4lg}.

We experiment with three encoders: BERT-base, T5-base, RoBERTa-base to investigate their efficacy on conditional tasks. We use the same encoder for $E_{diff}$ and $E_{cond}$. Our findings indicate that all three encoders demonstrate effectiveness across the tasks, achieving comparable performance levels.
However, no single encoder outperforms the others across all tasks. T5-base excels in question paraphrasing (QQP), while RoBERTa-base demonstrates superior performance on summarization (XSum) and on text simplification (Wiki-Auto) all encoders exhibit similar performance.

A comparison with other methods clearly shows that \textbf{TEncDM}, using any of the tested encoder models, outperforms popular diffusion embedding-based approaches. Furthermore, \textbf{TEncDM} with an optimal encoder achieves comparable performance to three AR baselines on QQP and XSum and outperforms them on Wiki-Auto.
\section{Conclusion}

In this work, we explore key details of the diffusion pipeline for text generation. We propose \textbf{TEncDM} which trains the diffusion model in the latent space of the language model encoding. Unlike embeddings, they contain contextual information which helps diffusion model to recover latents. To improve text generation performance, we analyse the effect of self-conditioning and conclude that it increases the magnitude of the model's predictions, which allows to reduce the number of generation steps. In addition, we propose an efficient decoder and noise scheduler that boost the generation quality. Extensive ablation on ROCStories and Wikipedia datasets demonstrates the impact of the proposed design choices. Finally, \textbf{TEncDM} outperforms recent diffusion models and some classical autoregressive methods in downstream task experiments.
\section{Limitations}

There are three limitations that warrant further investigation. First, the quality of the model can be improved by training diffusion encoder, decoder and denoising model simultaneously. However, we avoid doing so in order to avoid overcomplicating the approach.
Second, the samples from the latent space have a high dimensionality that depends on the sequence length, so the training of our method slows down significantly as the length increases. This problem can probably be overcome by training the autoencoder to map a text into a fixed-dimensional latent space, which is a great direction for further research. 
Third, as different diffusion encoders works better for different tasks, it is necessary to find the best one for each task.

\section*{Acknowledgments}
The paper was prepared within the framework of the HSE University Basic Research Program and was supported in part through computational resources of HPC facilities at HSE University. 

\bibliography{aaai25}

\appendix
\section{Decoder for embedding-based model} \label{app::emb_decoders}

We show that our proposed decoder is robust not only for encoding-based diffusion model, but also for embedding-based one. In Table \ref{tab::emb_decoders}, we compare our decoder described in Section \ref{sec::model_setup} with the commonly used rounding to the closest embedding on ROCStories dataset. It is easy to see that our decoder improves the text quality according to \textbf{mauve}. Also, it hugely improves \textbf{memorization} and \textbf{diversity}. Low value of \textbf{perplexity} for the rounding method comes from the low diversity and it does not imply the high quality of the generated samples.

\begin{table}[h]
\centering
\begin{tabular}{l|cccc}
\hline
\textbf{Decoder} & \textbf{ppl} $\downarrow$ & \textbf{mem} $\downarrow$ & \textbf{div} $\uparrow$ & \textbf{mauve} $\uparrow$ \\
\hline
Rounding & $32.4_{.41}$ & $.437_{.007}$ & $.252_{.005}$ & $.421_{.043}$ \\
Transformer &  \\
\; + $Cor(z_0)$ & $48.9_{.36}$ & $.371_{.003}$ & $.324_{.002}$ & $.600_{.016}$ \\
\hline
\end{tabular}
\caption{Decoders for the BERT embedding-based model.}
\label{tab::emb_decoders}
\end{table}

\section{Corruption for decoder training}\label{app::decoders}

Decoder is trained to map the latents $\hat{z}_0$ generated by the diffusion into text. These latents might be inaccurate and the decoder must take this into account in order to produce the best possible text. Therefore, we make the training task harder for the decoder by corrupting the input latents $z_0$ in order to mimic an imprecision of $\hat{z}_0$.

In this section, we experiment with two corruption techniques:
\begin{enumerate}
\item Replacing $z_0$ with $z_t$ by the diffusion forward process, $Cor(z_0) = \sqrt{\alpha_t} z_0 + \sqrt{1 - \alpha_t}\varepsilon = z_t$.
\item Adding a random Gaussian noise to decoder input, $Cor(z_0) = z_0 + \sigma \varepsilon$, where $\varepsilon \in \mathcal{N}(0, 1)$.
\end{enumerate}

\begin{figure}[H]
\begin{center}
\centerline{\includegraphics[width=\columnwidth]{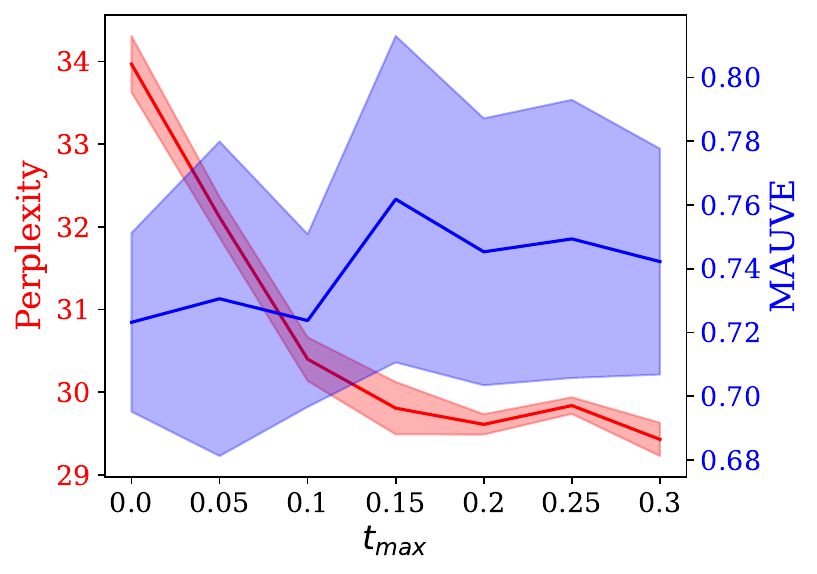}}
\caption{The dependence between the generation quality and the maximum amount of noise in $z_t$ during the decoder training.}
\label{fig::decoder_zt}
\end{center}
\end{figure}

\begin{figure}[H]
\begin{center}
\centerline{\includegraphics[width=\columnwidth]{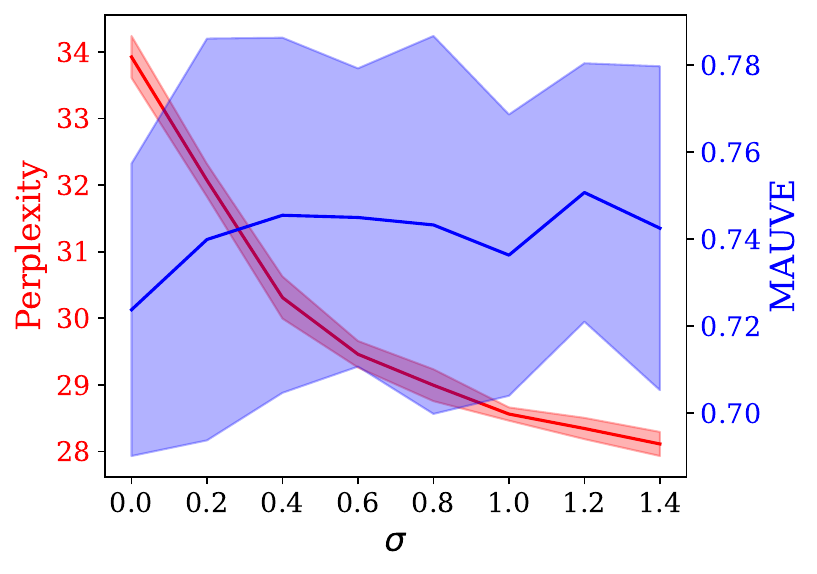}}
\caption{The dependence between the generation quality and the maximum amount of noise added to the latents during the decoder training.}
\label{fig::decoder_z0_noise}
\end{center}
\end{figure}

Both techniques introduce a random noise into the decoder input. However, the first one attempts to mimic samples from the diffusion model denoising trajectory. We implement it by randomly sampling a timestep from the range $t \in [0, t_{max}]$ and calculating the corresponding $z_t$. In Figure \ref{fig::decoder_zt}, we show the text generation quality in terms of Perplexity and MAUVE Score with respect to $t_{max}$. In Figure \ref{fig::decoder_z0_noise}, we present the similar result for the second decoder training technique with varying noise strength $\sigma$. To make the comparison fair we apply all decoders to the same latents produced by the diffusion model. Both plots suggest that there is an optimal amount of noise that should be added. However, the first technique results in a better performance.

\section{Self-conditioning}\label{app::self_conditioning}
\subsection{Self-conditioning increases prediction magnitude}

\begin{figure}[h]
\begin{center}
\centerline{\includegraphics[width=0.9\columnwidth]{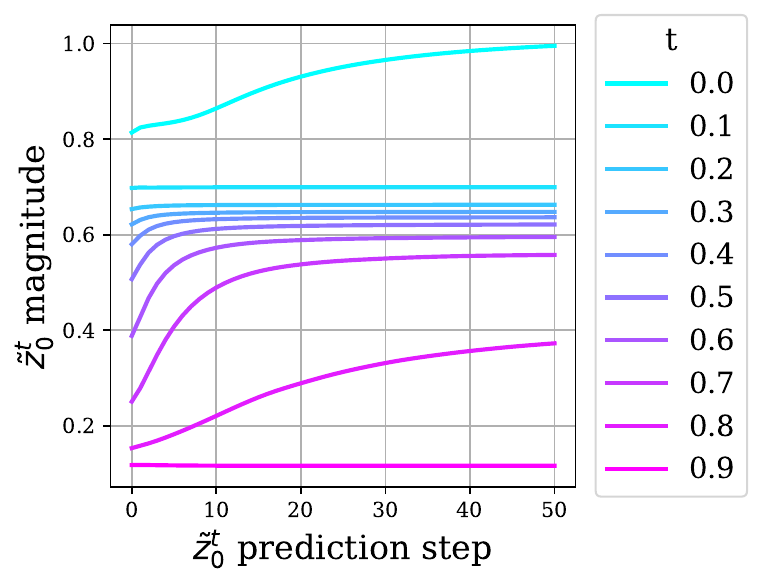}}
\caption{The effect of repeatedly predicting $\tilde{z}_0^t$ without deviating from the noisy latent $z_t$ on the magnitude of that prediction.}
\label{fig::repeated_sc_evaluation_vars}
\end{center}
\end{figure}

We show that self-conditioning tend to increase the magnitude of values of model's output by conducting the following experiment on ROCStories dataset. We sample $z_t$ using the diffusion forward process and predict $\tilde{z}_0^t = \hat{z}_\theta(z_t, t, \tilde{z}_0^t)$ from it several times. Each time we feed the model its previous prediction and do not change $z_t$ and timestep $t$. In Figure \ref{fig::repeated_sc_evaluation_vars}, we plot the trajectories of prediction magnitude obtained by this repeated prediction scheme for different timesteps $t$. The results show that the prediction magnitude grows at each step, even though we change only the sample, which we provide to a model using the self-conditioning. This allows us to conclude that self-conditioning is indeed responsible for the increase in prediction magnitude, which is reflected in the inference behaviour of the model.

\subsection{Wikipedia self-conditioning analysis}
\begin{figure}[h!]
\begin{center}
\centerline{\includegraphics[width=1\columnwidth]{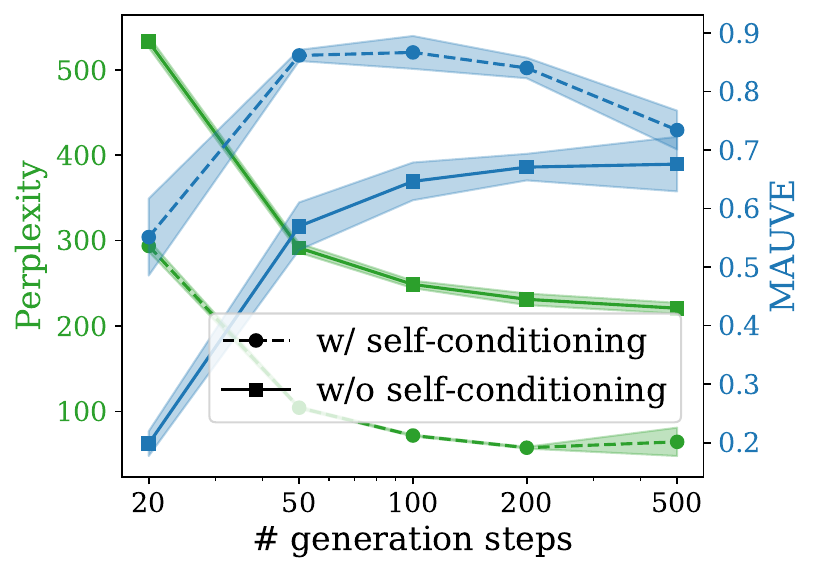}}
\caption{Comparison of models with and without self-conditioning on Wikipedia dataset.}
\label{fig::wiki_w_and_wo_sc}
\end{center}
\end{figure}

We conduct the experiments from Section \ref{sec::model_setup} on Wikipedia dataset to demonstrate the generalizability of self-conditioning effect. Figure \ref{fig::wiki_w_and_wo_sc} illustrates that with self-conditioning the MAUVE performance initially increases with the number of generation steps, reaching a certain level. Thereafter, it begins to decline. This behavior is similar to that observed on the ROCStories dataset. In contrast, for the model without self-conditioning, the quality exhibits a monotonic increase.
However, unlike the model trained on ROCStories, the Wikipedia-trained model achieves similar performance for both 50 and 100 generation steps.

\begin{figure}[h!]
\begin{center}
\centerline{\includegraphics[width=0.8\columnwidth]{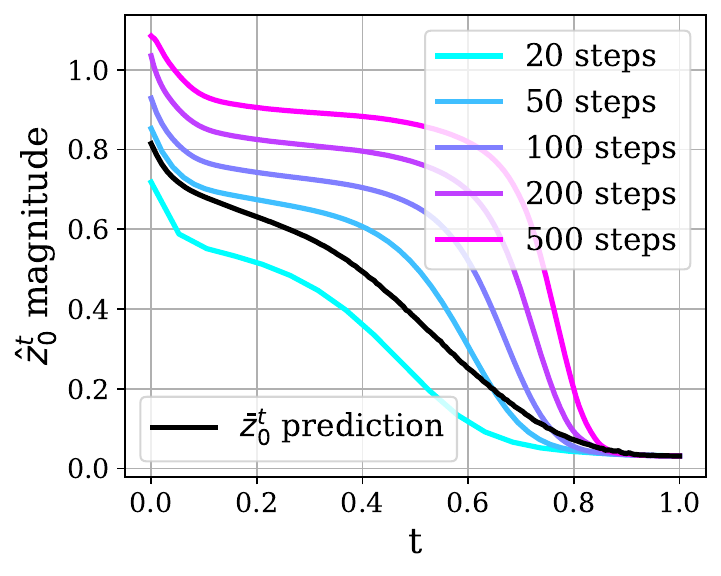}}
\caption{Comparison of prediction magnitudes for generation processes with different amount of steps on Wikipedia dataset.}
\label{fig::wiki_generation_z0_var_diff_steps}
\end{center}
\end{figure}

Figure \ref{fig::wiki_generation_z0_var_diff_steps} depicts the trajectories of prediction magnitudes for varying amounts of generation steps for the Wikipedia dataset. It can be observed that the trajectories exhibit a similar trend to those observed for the ROCStories on Figure \ref{fig::z0_var_steps}, although the spread of final prediction magnitudes is larger.

\section{Noise Schedulers}\label{app::noise_schedulers}

\begin{figure}[h!]
\begin{center}
\centerline{\includegraphics[width=0.75\columnwidth]{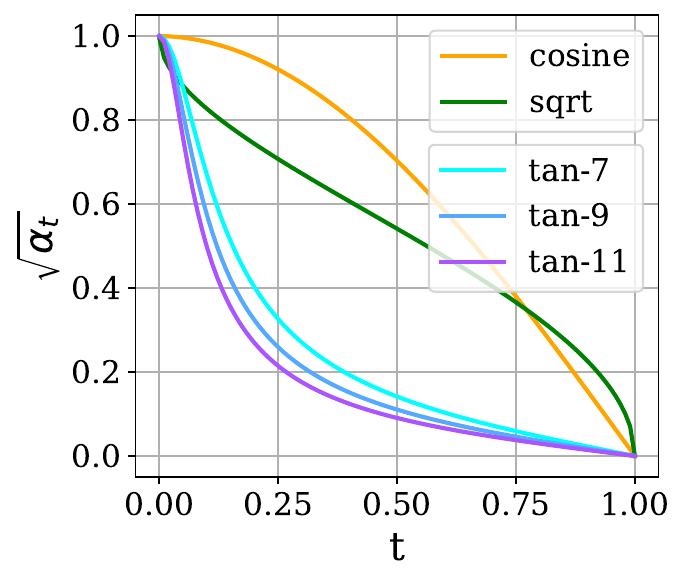}}
\caption{Visualizing different noise schedulers $\sqrt{\alpha_t}$.}
\label{fig::alphas}
\end{center}
\end{figure}

In Figure \ref{fig::alphas}, we visualize the noise addition rate in the forward process in terms of $\sqrt{\alpha_t}$ for different noise schedulers.

\section{Implementation details}\label{app::details}

We train our models using 4 A100 GPUs. The training takes approximately 6 hours for \textbf{ROCStories}, 15 hours for \textbf{Wikipedia}, 1 hour for \textbf{QQP}, 6 hours for \textbf{XSum} and 6 hours for \textbf{Wiki-Auto}. Table \ref{tab::details} shows hyperparameters for each dataset. For each dataset we chose the largest learning rate which keeps the training stable and train the model until convergence. The batch size for XSum was chosen smaller to fit on the GPU. We do not tune other hyperparameters, even though it might improve quality.

\begin{table*}[h]
    \centering
    \begin{tabular}{l|ccccc}
    \hline
    
    & \textbf{ROCStories} 
    & \textbf{Wikipedia} 
    & \textbf{XSum} 
    & \textbf{QQP}
    & \textbf{Wiki-Auto}
    \\
    
\hline
    Diffusion Trainable Params       & \multicolumn{5}{c}{101M} \\
    Transformer Layers               & \multicolumn{5}{c}{12} \\
    Transformer Dim                  & \multicolumn{5}{c}{768} \\
    Self-Attention Heads             & \multicolumn{5}{c}{12} \\
    Decoder Trainable Params         & \multicolumn{5}{c}{44M} \\
    Decoder Self-Attention Heads     & \multicolumn{5}{c}{16} \\
    Decoder Transformer Layers       & \multicolumn{5}{c}{3} \\
    Optimizer                        & \multicolumn{5}{c}{AdamW} \\
    Learning Rate                    & 2e-4 & 2e-4 & 2e-4 & 4e-4 & 4e-4 \\
    ($\beta_1$, $\beta_2$)           & \multicolumn{5}{c}{(0.9, 0.980)} \\
    Warmup Steps                     & \multicolumn{5}{c}{5000} \\
    Learning Rate Sch                & \multicolumn{5}{c}{Constant} \\
    Weight Decay                     & \multicolumn{5}{c}{0.01} \\ 
    Gradient Clipping                & \multicolumn{5}{c}{1} \\
    EMA Decay                        & \multicolumn{5}{c}{0.9999} \\
    Batch Size                       & 512 & 512 & 128 & 512 & 512  \\
    Training Steps                   & 200k & 500k & 100k & 200k & 200k \\
    Max Seq Length                   & 80  & 128   & 64 & 50 & 64 \\
    Max Context Length               & –  & –  & 512 & 50 & 64 \\
    Sampling steps                   & \multicolumn{5}{c}{50} \\
    Guidance coefficient             & –  & –  & 0.5 & 0.5 & 0.5 \\
\hline
\end{tabular}
\caption{Hyperparameter values for TEncDM across different datasets.}
\label{tab::details}
\end{table*}

\section{Dataset Statistics}\label{app::datasets}

\paragraph{ROCStories} The dataset consists of 98,161 instances. 93,161 instances are held out for training, 1,000 instances for validation, 4,000 instances for testing.

\paragraph{Wikipedia} For large-scale experiments, we utilize the English Wikipedia subset from the ROOTS corpus \cite{bigscience}. Additionally articles containing fewer than 600 characters are removed, and the remaining articles are segmented into sequences of approximately 128 tokens each.
The dataset comprises a total of 15.4M sequences, of which 10,000 are allocated as a validation set. 

\paragraph{XSum} This dataset is used for summarization task and it contains 204k BBC articles, which are provided as document and summary pairs and covered wide range of topics (Sports, Politics, etc.). It has 204,045 training instances, 11,332 validation instances, and 11,334 test instances.

\paragraph{QQP} The subset of QQP dataset, proposed in \cite{diffuseq}, consists of 144k question pairs from the Quora platform that are paraphrases of each other. It has 144,715 training instances, 2,048 validation instances, and 2,500 test instances.

\paragraph{Wiki-Auto} The dataset proposed by \cite{wiki_auto} is designed for solving the text simplification task. The goal is to turn complex text into sequences with simplified grammatical structures and a more limited range of vocabulary.
We use the preprocessing scheme outlined by \cite{diffuseq}, resulting in a dataset with 677,000 training instances, 2,050 validation instances, and 5,000 test instances.

\section{Low MAUVE for RoBERTa encoder}\label{app::mauve_roberta}

Table \ref{tab::encoders} illustrates that, despite satisfactory perplexity, memorialization, and diversity values, the MAUVE metric exhibits a low value for RoBERTa encoder. This outcome is not attributable to inferior text quality but largely to the discrepancy in the lengths of the generated and validation texts. MAUVE is estimated based on GPT2 encoding of the text, which extracts information about text length, among other elements. To validate this hypothesis, two experiments were conducted. 

\begin{figure}[h!]
\begin{center}
\centerline{\includegraphics[width=1\columnwidth]{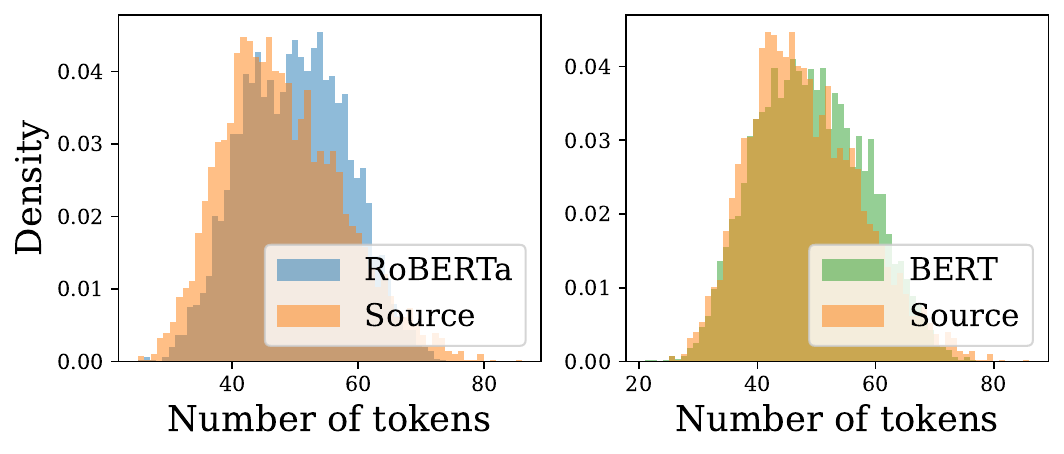}}
\caption{Text length distribution for source texts and TEncDM generation with RoBERTa and BERT encoders.}
\label{fig::roberta_bert_hists}
\end{center}
\end{figure}

\paragraph{Experiment 1}
In Figure \ref{fig::roberta_bert_hists}, we visualize the text length distributions for TEncDM with RoBERTa and BERT encoders, and for the source text. It is clear that the BERT's distribution is much more similar to the source text distribution than RoBERTa's. This suggests that the reason for drop in MAUVE value might indeed be the text length.

\paragraph{Experiment 2}

To match the length distributions, we truncate all generated and source texts to 40 tokens and measure MAUVE for them. The measurement results are summarized in Table \ref{tab::truncated_mauve}. The obtained values increased significantly compared to the values in Table 1, and also became closer to each other. This indicates that MAUVE is indeed dependent on the length of texts.

\begin{table}[H]
\centering
\begin{tabular}{l|c}
\hline
\textbf{Encoder} & \textbf{mauve} $\uparrow$ \\
\hline
BERT & $.940_{.013}$ \\
RoBERTa & $.913_{.009}$ \\
\hline
\end{tabular}
\caption{MAUVE for TEncDM with BERT and RoBERTa encoders on texts truncated to 40 tokens.}
\label{tab::truncated_mauve}
\end{table}

\begin{figure}[h!]
\begin{center}
\centerline{\includegraphics[width=0.8\columnwidth]{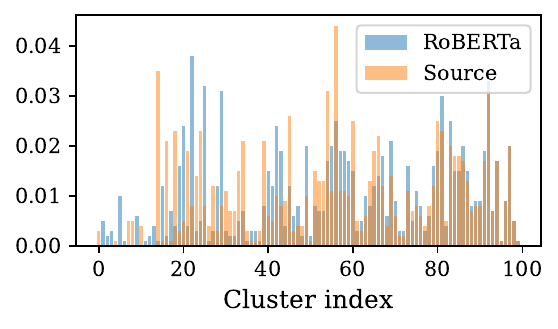}}
\caption{Histograms used for MAUVE calculation.}
\label{fig::roberta_mauve_hist}
\end{center}
\end{figure}

\paragraph{Experiment 3}
According to the MAUVE calculation algorithm, we first cluster vector representations of texts using K-Means. Then for both generated and source texts we construct the histogram of cluster sizes based on the resulting clustering. If the histograms are similar, it can be inferred that the generated and source texts have similar distribution, which indicates a high degree of generation quality. Conversely, if the histograms are different, it can be concluded that the distribution of generated texts is distinct from the distribution of source texts. Such histograms for RoBERTa encoder are shown in Figure \ref{fig::roberta_mauve_hist}.

We can analyze the differences in the distributions by comparing clusters, wherein the number of generated and real texts is balanced, with unbalanced clusters.

\begin{figure}
\begin{center}
\centerline{\includegraphics[width=0.8\columnwidth]{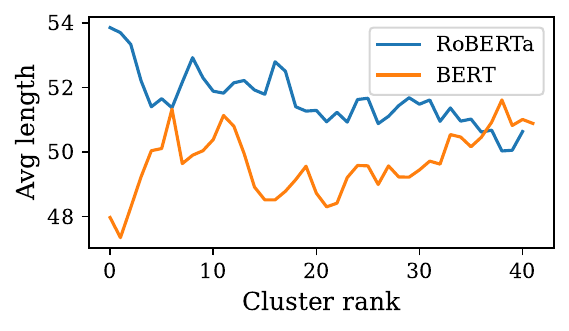}}
\caption{Smoothed average text lengths for clusters sorted by entropy.}
\label{fig::roberta_bert_lens}
\end{center}
\end{figure}

First, we calculate the entropy of the distribution of generated and real texts in each cluster to measure its balance. The resulting entropies are then sorted to yield a ranked list of clusters. Then, as we hypothesise that the main difference between texts is their length, we calculate the average text length in each cluster in terms of tokens. We take in consideration only those clusters, in which there are more generated texts than source ones. We plot these lengths sorted by cluster entropy in Figure \ref{fig::roberta_bert_lens} for both RoBERTa and BERT encoders. We have applied smoothing to each plot in order to enhance the visibility of the trend.

In Figure \ref{fig::roberta_bert_lens}, the entropy increases from left to right with the cluster rank, which means that the clusters on the left are more imbalanced than those on the right. We observe that the length of RoBERTa's texts decreases with an increase in balance, whereas the length of BERT's texts remains relatively constant. This suggests that RoBERTa clusters without source texts tend to have longer texts, which is the underlying reason for the decline in MAUVE performance.

\section{Training time comparison}

In Table \ref{tab::time_comparison} we provide time measurements for one training iteration of the proposed method and its modifications: without diffusion encoder (embedding-based diffusion model) and without self-conditioning. We performed the measurements on a single Nvidia A100 GPU for the batch of 512 texts with a sequence length of 128. We notice that the addition of the diffusion encoder slows down the training by only 15\% and the addition of self-conditioning by 20\%. 
Note that the generation speed is the same for models with and without diffusion encoder, as the diffusion encoder is not used during generation. 

\begin{table}[H]
\centering
\begin{tabular}{l|c}
\hline
\textbf{Model} & \textbf{Time for one iteration (s)} \\
\hline
TEncDM & $.730_{.147}$ \\
\, w/o $E_{diff}$ & $.634_{.134}$ \\
\, w/o self-conditioning & $.605_{.110}$\\
\hline
\end{tabular}
\caption{Training time comparison for TEncDM and its modifications.}
\label{tab::time_comparison}
\end{table}

\section{Scalability}

We investigate the scalability of TEncDM.
Our model leverages both a text encoder and a diffusion model, offering two avenues for parameter expansion: scaling the diffusion model alone or scaling both the encoder and diffusion model concurrently. 
In this section, we focus on the first approach, where we exclusively increase the size of the diffusion model. 
Specifically, we increased the number of transformer heads from 12 to 16, while maintaining the size of each head, and doubled the number of layers from 12 to 24, bringing the total number of parameters to 223M. We call the standard model ``base'' and the increased model ``large''.
In Table \ref{tab::scaling} we show comparison of base and large diffusion models on \textbf{Wikipedia} dataset. The results demonstrate the effectiveness of the scaling strategy.
The second approach, involving simultaneous scaling of both the encoder and diffusion model, necessitates significantly greater computational resources and training time.
Therefore, we defer the exploration of this approach to future research.

\begin{table}
\centering
\begin{tabular}{l|llll}
\hline
\multicolumn{1}{l|}{\textbf{Model size}}
    & \multicolumn{1}{c}{\textbf{ppl} $\downarrow$} 
    & \multicolumn{1}{c}{\textbf{mem} $\downarrow$}
    & \multicolumn{1}{c}{\textbf{div} $\uparrow$}
    & \multicolumn{1}{c}{\textbf{mauve} $\uparrow$}\\
\hline
base (105M) & $104.4_{2.1}$ & $.286_{.002}$ & $.504_{.003}$ & $.874_{.011}$ \\
large (223M) & $84.5_{0.8}$  & $.298_{.001}$ & $.505_{.002}$ & $.909_{.011}$ \\
\hline
\end{tabular}
\caption{Scalability of the diffusion model on Wikipedia.}
\label{tab::scaling}
\end{table}

\begin{table*}
\centering
\begin{tabular}{l | p{0.8\linewidth}}
\hline
\vtop{\hbox{\strut BERT embeddings}\hbox{\strut with}\hbox{\strut Transformer decoder}} & \;\; Last week my brother brought my skateboard with me. He started using the skateboard after half an hour long. I \hl{bit} my leg and started to \hl{fall out of my foot}. My brother got into the piece. He was able to \hl{scolded} me and take me to the hospital. \\
& \;\; Liz was in the kitchen \hl{watching watching} TV. She heard a sharp s Henk. She picked it up and ran downstairs to grab what her sandwich was. She quickly grabbed a hot cheese from her sandwich. She put the sandwich on the stove and turned it \hl{down the plate}. \\
& \;\;Larry and his girlfriend were making family dinner last night. After a long time, they decided to make lasagna. They made the meat mix and tested the bread. They had to cut \hl{the meat off the pizza}. It \hl{lit up} as soon as it was done.\\
\hline
\vtop{\hbox{\strut BERT encodings}\hbox{\strut with}\hbox{\strut Transformer decoder}} & \;\; Emily wanted her nails become pink. She took some \hl{nailolish} from a grocery store and thought it looked horrible. She tried everything to get rid of it. It ended up making a ton of mess. Emily had to throw the mess all out.\\
& \;\; Bianca was at a local tennis party. She was having a good time with her friends. Suddenly she realized that she had lost her wallet! She searched for an hour to no avail. Luckily she found it there and was glad that she didn't lose it. \\
& \;\; Ally wakes up one morning feeling very well. Ally realizes she has a pregnancy test. Ally decides she will go to the doctor to get her test. Ally is shocked when the results show that she is pregnant. Ally is very excited when her pregnancy test \hl{is confirmed}. \\
\hline
\end{tabular}
\caption{Examples of generated texts for different models on ROCStories dataset. Generation inaccuracies are highlighted.}
\label{tab::generated_examples}
\end{table*}

\section{Comparison with GPT-2}\label{app::comparison_gpt2}

We compare TEncDM with fine-tuned GPT-2-small \cite{gpt2} on an unconditional generation task using \textbf{ROCStories} and \textbf{Wikipedia} datasets. We use the Nucleus sampling with $p = 0.95$ for the GPT-2 generation, as it produced the best results. Both models have similar amount of parameters (124M for GPT-2 and 145M for TEncDM). The result of the comparison is presented in Table \ref{tab::rocstories_gpt2} and it shows that our approach suprasses fine-tuned GPT-2-small by all metrics except perplexity, because GPT-2 tends to memorise the training data set more and has a lower diversity. However, the perplexity comparison is unfair as it is computed with the GPT-2-large model, which behaves similarly to GPT-2-small.

\begin{table}[H]
\centering
\small
\begin{tabular}{l|cccc}
\hline
\textbf{Decoder} & \textbf{ppl} $\downarrow$ & \textbf{mem} $\downarrow$ & \textbf{div} $\uparrow$ & \textbf{mauve} $\uparrow$ \\
\hline
\multicolumn{5}{c}{\textbf{ROCStories}} \\
\hline
GPT2-small FT & $\textbf{15.5}_{.11}$ & $.519_{.004}$ & $.269_{.003}$ & $.739_{.031}$ \\
TEncDM & $29.1_{.89}$ & ${\textbf{.453}}_{.003}$ & ${\textbf{.295}}_{.002}$ & $\textbf{.762}_{.043}$ \\
\hline
\multicolumn{5}{c}{\textbf{Wikipedia}} \\
\hline
GPT2-small FT & $\textbf{15.3}_{.18}$ & $.508_{.004}$ & $.380_{.003}$ & $.708_{.025}$ \\
TEncDM & $104.4_{2.1}$ & $\textbf{.286}_{.002}$ & $\textbf{.504}_{.003}$ & $\textbf{.874}_{.011}$ \\
\hline
\end{tabular}
\caption{Comparison with GPT-2 on unconditional generation.}
\label{tab::rocstories_gpt2}
\end{table}

\section{Generation examples}\label{app::generated_examples}

We show generation examples for diffusion model trained in the latent space of embeddings and encodings in Table \ref{tab::generated_examples}.

\end{document}